\documentclass{article}

\usepackage{microtype}
\usepackage{graphicx}
\usepackage{subfigure}
\usepackage{booktabs} 
\usepackage{amsmath} 
\usepackage{amssymb}
\usepackage{hhline}
\usepackage{textcomp}

\usepackage[hyphens]{url}
\usepackage{breakurl}
\usepackage[breaklinks]{hyperref}
\usepackage{listings}
\usepackage{tabularx}

\usepackage{xcolor}
\definecolor{codegreen}{rgb}{0,0.6,0}
\definecolor{codegray}{rgb}{0.5,0.5,0.5}
\definecolor{codepurple}{rgb}{0.58,0,0.82}
\definecolor{backcolour}{rgb}{0.95,0.95,0.92}
\usepackage[accepted]{icml2019}

\begin{document}

\twocolumn[
\title{optimizn: a Python Library for Developing Customized Optimization Algorithms}
\date{\vspace{-0.2in}}
\maketitle



\icmlsetsymbol{equal}{*}

\begin{icmlauthorlist}
\icmlauthor{Akshay Sathiya, Azure Core Insights Data Science, asathiya@microsoft.com}{equal}
\icmlauthor{Rohit Pandey, Azure Core Insights Data Science, ropandey@microsoft.com}{equal}
\end{icmlauthorlist}
\vspace{0.4in}
\begin{abstract}
Combinatorial optimization problems are prevalent across a wide variety of domains. These problems are often nuanced, their optimal solutions might not be efficiently obtainable, and they may require lots of time and compute resources to solve (they are NP-hard). It follows that the best course of action for solving these problems is to use general optimization algorithm paradigms to quickly and easily develop algorithms that are customized to these problems and can produce good solutions in a reasonable amount of time. In this paper, we present optimizn, a Python library for developing customized optimization algorithms under general optimization algorithm paradigms (simulated annealing, branch and bound). Additionally, optimizn offers continuous training, with which users can run their algorithms on a regular cadence, retain the salient aspects of previous runs, and use them in subsequent runs to potentially produce solutions that get closer and closer to optimality. An earlier version of this paper was peer reviewed and published internally at Microsoft. 
\\

\end{abstract}
\vspace{0.4in}
]

\printAffiliationsAndNotice{\icmlEqualContribution} 
\clearpage

\section{Introduction}
There are a myriad of combinatorial optimization problems that arise in cloud computing at Microsoft Azure and in virtually any other industry or line of work. Many optimization problems are $NP$-hard, so their optimal solutions are unlikely to be obtainable efficiently/in polynomial time. Even if the problems are not $NP$-hard, they are often very nuanced, making it impractical to develop several problem-specific algorithms from scratch.

The solutions to these problems cannot be generated trivially, without risking undesirable business impacts. For instance, consider the environment design problem in Azure \cite{EnvDesign}, where testing environments are designed to catch regressions in Azure's internal programs before deployment. Trivially generated testing configurations could cause certain regressions/errors/bugs to be missed in pre-production testing, be deployed to production, and impact customers. Hence, it is important to develop algorithms that find satisfactorily near-optimal solutions (``good" solutions), in a reasonable amount of time. 

This paper presents optimizn, a code library created by the authors, which can be used to develop optimization algorithms that are customized to specific optimization problems and can quickly produce good solutions to those problems. 

optimizn offers simulated annealing and branch and bound, with continuous training for both. Simulated annealing and branch and bound are general optimization algorithm paradigms that can be customized to specific optimization problems. Continuous training allows both algorithms to run, save their problem parameters, best solution found, and state, and resume running from that state later. This allows the algorithms to find good solutions in situations where compute resources and time are only available in disjoint intervals. This is particularly helpful when the problems are relatively slow moving and static (for example, when the horizon of data collected is a long rolling window like 30 days).

optimizn can solve a variety of optimization problems in practice, inside and outside of Azure. It has already been used to solve the environment design problem in Azure. The environment design problem and the system that solves it (which uses optimizn) are discussed in greater detail in other literature written by the authors of this paper \cite{EnvDesign}.

\section{Background Information}

\subsection{$NP$-Hard Problems}

$NP$-hard problems \cite{CLR4} are decision problems, which need not be in $NP$, that all problems in $NP$ are reducible to in polynomial time.
$NP$-hard problems outside of $NP$ are also outside of $P$, so they cannot be solved in polynomial time. At the time of this writing, no polynomial time algorithms have been found for any $NP$-hard problem in $NP$ ($NP$-complete problem), so $NP$-hard problems in $NP$ are considered unlikely to be solvable in polynomial time. It follows that if a problem is $NP$-hard, it is unlikely to be solvable in polynomial time.

$P/NP$ complexity theory mainly pertains to decision problems, but can be applied to optimization problems by bounding the optimality of their solutions \cite{CLR4}. Optimization problems are at least as hard as their corresponding decision problems, so an optimization problem is $NP$-hard (consequently, its optimal solution is unlikely to be obtainable in polynomial time) if its corresponding decision problem is $NP$-hard.

\subsection{Simulated Annealing}

Simulated annealing \cite{SimAnnealOverview} is inspired by the annealing of solids, where heated metals in a malleable state are cooled into a desired state. In simulated annealing, a solution is iteratively modified into a more optimal solution (analogous to cooling a solid). To escape local minima of the cost/objective function, simulated annealing features random restarts and occasionally (based on a temperature value, which decreases over the iterations) allows modifications that produce a less optimal solution.

\subsection{Branch and Bound}

Branch and bound \cite{BranchAndBound1} represents the problem space as a tree, where the root node is the original constrained optimization problem and its descendant nodes are more constrained versions of the problem, which have new constraints in addition to the constraints of their ancestor nodes. These are additional constraints introduced by the algorithm, distinct from the constraints of the original problem. Non-leaf nodes are partially constrained versions of the problem (with multiple solutions) and correspond to partial solutions. Leaf nodes are fully constrained versions of the problem (with one solution) and correspond to complete solutions. The tree is grown by iteratively branching on one of its nodes to get more nodes, which correspond to more constrained versions of the problem and solutions that are closer to completion. As the tree is grown and nodes are evaluated, nodes (and their subtrees) are pruned if their lower bounds indicate they will not lead to a solution more optimal than the most optimal solution seen so far.

Since the leaf nodes may take a while to reach, partial solutions can be completed in some way (specific to the problem) to yield complete solutions faster and prune subtrees earlier. This completion of partial solutions is based on \cite{BranchAndBound2}. In this paper, ``traditional" and ``look-ahead" branch and bound refer to branch and bound without and with the completion of partial solutions, respectively.

\section{Related Works}

\subsection{Simulated Annealing}

The following software packages for simulated annealing have been reviewed: \texttt{optim\_sa} in R \cite{OptimSAR}, \texttt{anneal} in SciPy \cite{SciPyAnneal}, \texttt{dual\_annealing} in SciPy \cite{SciPyDualAnnealing}, and \texttt{basinhopping} in SciPy \cite{SciPyBasinHopping}.

Implementations of simulated annealing in statistical languages like R or Python's SciPy apply to problems where the inputs are arrays with continuous elements. This is despite the SciPy documentation noting: ``In practice [simulated annealing] has been more useful in discrete optimization than continuous optimization, as there are usually better algorithms for continuous optimization problems" \cite{SciPyAnneal}. So, even the traveling salesman problem, which is often used to demonstrate the effectiveness of simulated annealing, cannot be solved with these libraries since it is a combinatorial optimization problem with non-continuous inputs.

Adding an additional layer of complexity, many optimization problems that are present in practice are combinatorial constrained optimization problems, needing further customization to take the constraints into account.

This calls for a library that can perform simulated annealing on all kinds of combinatorial optimization problems (constrained and unconstrained). This is where optimizn shines and better alternatives are likely not available, even if using optimizn is slightly more involved and more responsibility lies with the user than other libraries/packages.

Hence, the optimizn library offers heuristics for solving NP-hard problems (like simulated annealing), where the customizable components of the algorithm can be implemented by the user for their specific optimization problem, regardless of the nature of their inputs (discrete/continuous).

\subsection{Branch and Bound}

The following software packages for branch and bound have been reviewed: IBM$^\circledR$ Decision Optimization CPLEX$^\circledR$ \cite{CPLEX} (referred to simply as ``CPLEX"), PyBnB \cite{PyBnB}.

CPLEX uses branch and bound to solve constrained optimization problems formulated in a mixed-integer or integer programming context \cite{CPLEXBnB} (i.e. with numerically valued variables and mathematical expressions and equations/inequalities). However, some optimization problems, like the environment design problem, are better expressed in other ways that allow optimization algorithms to be more efficient with time and memory \cite{EnvDesign}. Other optimization problems may follow a similar story, which motivates the need for a library with a more general approach to branch and bound, even if using the library is more involved.

The PyBnB package offers a general branch and bound implementation, where the user implements methods based on the specifics of their optimization problem. Additionally, PyBnB allows the user to specify how their problem state can be saved and loaded so execution of their branch and bound algorithm can be resumed from where a previous execution left off (this is essentially continuous training). 

However, PyBnB does not support the completion of partial solutions, which is undesirable in practice since finding complete solutions (reaching the leaf nodes) may take time. If partial solutions could be completed, then complete solutions could be found quicker (before reaching the leaf nodes) and subtrees can be pruned earlier.

Additionally, PyBnB does not accept an initial solution. For problems where an initial solution is available/obtainable, this is undesirable in practice since time is wasted on subtrees that will not yield a solution more optimal than the initial solution. If an initial solution could be provided, those subtrees could be pruned.

Hence, optimizn also offers branch and bound, where the customizable components of the algorithm can be implemented by the user for their specific optimization problem. The user can also complete partial solutions and provide an initial solution (both are optional) to improve the algorithm's performance.

\section{Code and Contracts}

There are three main classes in the optimizn library: \texttt{OptProblem} (pertains to all optimization problems and continuous training), \texttt{SimAnnealProblem} (pertains to optimization problems solved using simulated annealing), and \texttt{BnBProblem} (pertains to optimization problems solved using branch and bound).

To implement an optimization algorithm, the user creates a class for their optimization problem (referred to as the ``optimization problem class") which is a subclass of either the \texttt{SimAnnealProblem} or \texttt{BnBProblem} class, based on the optimization algorithm being implemented (simulated annealing or branch and bound, respectively). Both the \texttt{SimAnnealProblem} and \texttt{BnBProblem} classes are subclasses of the \texttt{OptProblem} class.

The optimization problem class inherits a solver method that executes the optimization algorithm and a persist method to save the problem parameters and instance of the optimization problem class (which contains the current state of the algorithm, the most optimal solution found, and the cost of that solution) for future use/reference (continuous training). The optimization problem class also inherits other methods that correspond to the customizable components of the optimization algorithm. These methods are used in the solver method, and must be implemented by the user (as needed for their specific optimization problem, taking the constraints into account).

To run their optimization algorithm, the user can simply instantiate their optimization problem class (or load a saved instance) and call the solver method. The result of the algorithm is the best/most optimal solution found and its cost (the values of the \texttt{best\_solution} and \texttt{best\_cost} attributes, respectively). After the run has finished, the user can call the persist method to save the problem parameters and instance of the optimization problem class for continuous training.

The code and contracts for the optimizn library's simulated annealing, branch and bound, and continuous training offerings are shown in the following subsections.

\subsection{Optimization Problem Class with Continuous Training}

The \texttt{OptProblem} class contains logic required for all optimization problems and continuous training.

The following methods must be implemented by the user in their optimization problem class.
\begin{itemize}
\item \texttt{get\_initial\_solution}: Produces the initial solution (optional for branch and bound).
\item \texttt{cost}: Computes the value of the objective function (cost) for a given solution.
\item \texttt{cost\_delta}: Computes the difference between two cost values (defaults to the difference between the first and second cost values, can be overridden).
\item \texttt{persist}: Saves the optimization problem parameters and optimization problem class instance. The default implementation (which uses the \texttt{pickle} library) can be overridden by the user.
\end{itemize}

To perform continuous training, the user must call the \texttt{persist} method after running their optimization algorithm. To resume the run later, the user can load the saved optimization problem parameters and compare them to the current parameters. If they match, the user can load the saved optimization problem class instance and call the solver method. For the default \texttt{persist} method, the \texttt{load\_latest\_pckl} function can be used to load the saved parameters and optimization problem class instance.

\subsection{Simulated Annealing Class}

The \texttt{SimAnnealProblem} class is based on \cite{SimAnnealOverview, SimulatedAnnealing1, SimulatedAnnealing2}. The \texttt{SimAnnealProblem} class extends the \texttt{OptProblem} class and contains the logic for simulated annealing.

The following methods must be implemented by the user in their optimization problem class.
\begin{itemize}
\item \texttt{next\_candidate}: Produces a new solution by modifying the current solution.
\item \texttt{reset\_candidate}: Produces a new solution that becomes the current solution under the specified reset probability. Used for performing random restarts. Defaults to \texttt{get\_initial\_solution}, can be overridden.
\item \texttt{get\_temperature}: Gets the temperature given the number of iterations since the last random restart. Defaults to the below function \cite{SimulatedAnnealing1, SimulatedAnnealing2}, can be overridden.

$$f(x) = \frac{4000}{1 + e^{x / 3000}}$$
\end{itemize}

The simulated annealing algorithm is performed in the \texttt{anneal} method, which uses the methods implemented by the user in the optimization problem class. The user can specify the number of iterations, the reset probability, and the time limit (in seconds) through the \texttt{n\_iters}, \texttt{reset\_p}, and \texttt{time\_limit} arguments (respectively) of the \texttt{anneal} method.

\subsection{Branch and Bound Class}

The \texttt{BnBProblem} class is based on \cite{BranchAndBound1, BranchAndBound2}. The \texttt{BnBProblem} class extends the \texttt{OptProblem} class and contains the logic for branch and bound. 

Through the \texttt{bnb\_selection\_strategy} argument in the \texttt{BnBProblem} class constructor, the user can specify the strategy for selecting the next node in the tree to evaluate. The supported selection strategies are depth-first (selects and evaluates nodes in a depth-first-search manner), depth-first-best-first (selects and evaluates nodes in a depth-first-search manner, prioritizes lower bound for nodes of the same depth in the tree), or best-first-depth-first (selects and evaluates the node with the lowest lower bound, prioritizes depth in tree for nodes with the same lower bound).

The space complexity (maximum number of nodes in the tree) under the depth-first-best-first selection strategy is polynomially bounded with respect to the maximum branch factor and maximum depth of the tree, and is exponentially bounded with respect to the maximum depth of the tree \cite{BranchAndBound1}. The space complexity under the depth-first selection strategy is linearly bounded with respect to the maximum depth of the tree, since every time a node is branched on, the first solution yielded from the branching is evaluated and branched on before the next solution is yielded, evaluated, and branched on.

The following methods must be implemented by the user in their optimization problem class.
\begin{itemize}
\item \texttt{get\_root}: Produces the root solution, a partial solution from which other partial solutions and complete solutions are obtainable through branching. The root solution corresponds to the root node of the tree.
\item \texttt{is\_feasible}: Determines if a given partial/complete solution is feasible (i.e., is a complete solution that adheres to the constraints of the problem).
\item \texttt{branch}: Grows the tree from a given partial solution by generating ``child" solutions have additional properties (are closer to completion) and correspond to more constrained versions of the problem.
\item \texttt{lbound}: Computes a lower bound on the cost of all feasible solutions obtainable through branching on a given partial solution (i.e., all feasible solutions in the given partial solution's subtree).
\item \texttt{complete\_solution}: Completes a given partial solution (only needed for look-ahead branch and bound).
\end{itemize}

The branch and bound algorithm is performed in the \texttt{solve} method using the methods implemented by the user in the optimization problem class. The user can specify the number of iterations, the time limit (in seconds), and the type of branch and bound algorithm they would like to execute (traditional or look-ahead), through the \texttt{iters\_limit}, \texttt{time\_limit}, and \texttt{bnb\_type} arguments (respectively) of the \texttt{solve} method.

\section{Applicability to a Well-Known $NP$-Hard Problem}

The optimizn library can be used on many $NP$-hard optimization problems to quickly obtain good solutions. Let's see this in action for a very popular optimization problem, the symmetric traveling salesman problem \cite{BranchAndBound1} (simply referred to as the ``traveling salesman problem" in this paper), which is an $NP$-hard problem \cite{CLR4}.

\subsection{Traveling salesman problem}

In this paper, the traveling salesman problem is formulated as follows.

Inputs:
\begin{itemize}
\item Symmetric adjacency matrix representing a graph of $n$ cities: $A \in \mathbb{R}^{+, n \times n}$, where $A_{i, j} = A_{j, i}$ $\forall i, j \in \{1, 2, ..., n\}$
\end{itemize}

Outputs:
\begin{itemize}
\item Path of cities $s = [s_1, s_2, ..., s_n]$
\end{itemize}

Constraints:
\begin{itemize}
\item Solution path is of length $n$, $|s| = n$
\item Every city is covered, $\forall i \in \{1, 2, \dots, n\}, i \in s$
\end{itemize}

Optimization:
\begin{itemize}
\item Minimize the length of the path (including distance to start city), $\Sigma_{i=1}^{n-1}A_{s_i, s_{i + 1}} + A_{s_n, s_1}$
\item Optimal solution is $\text{min}_s \{ \Sigma_{i=1}^{n-1}A_{s_i, s_{i + 1}} + A_{s_n, s_1} \}$
\end{itemize}

\subsection{Simulated Annealing Algorithm}

The optimizn library's simulated annealing algorithm for the traveling salesman problem is based on \cite{SimulatedAnnealing1, SimulatedAnnealing2}.

The following methods inherited from the \texttt{OptProblem} and \texttt{SimAnnealProblem} classes have been implemented.

\begin{itemize}
\itemsep 0em
\item \texttt{get\_initial\_solution}: Assembles an initial path with the cities in increasing, numerical order.
\item \texttt{reset\_candidate}: Randomly generates a path of cities.
\item \texttt{cost}: Sum of distances between each pair of adjacent cities in the path, and between the last and first cities. 
\item \texttt{next\_candidate}: Produces a new path by swapping the places of two cities in a given path.
\item \texttt{get\_temperature}: Implemented as the below function. 
$$f(x) = \frac{4000}{1 + e^{x / 10000}}$$
\end{itemize}
            
\subsection{Branch and Bound Algorithm}

The optimizn library was used to develop a branch and bound algorithm for the traveling salesman problem.

The following methods inherited from the \texttt{OptProblem} and \texttt{BnBProblem} classes have been implemented. 

\begin{itemize}
\itemsep 0em
\item \texttt{get\_initial\_solution}: Assembles an initial path with the cities in increasing, numerical order.
\item \texttt{get\_root}: Returns an empty path.  
\item \texttt{complete\_solution}: Appends unvisited cities to a given incomplete path, in a random order.
\item \texttt{cost}: Sum of distances between each pair of adjacent cities in the path, and between the last and first cities.
\item \texttt{branch}: Given a path, returns a list of new paths, each of which is the given path followed by an unvisited city.
\item \texttt{is\_feasible}: checks if the path length is equal to the number of cities and each city is visited only once.
\item \texttt{lbound}: takes the sum of the distance values corresponding to adjacent cities in the partial solution and the $k$ lowest remaining distance values between any two cities ($k$ is the number of distance values needed to visit the unvisited cities and return to the first city).
\end{itemize}

\section{Applicability to a Niche $NP$-Hard Problem}

A niche $NP$-hard problem encountered in practice is the environment design problem in Azure, which is formulated as follows \cite{EnvDesign}.

Inputs:
\begin{enumerate}
\item An undirected graph $G = (V = V_1 \cup \dots \cup V_d, E)$ where the vertices in $V$ are split into $d \geq 2$ groups, each group representing a testing dimension (e.g. HW model, VM type) and each vertex representing a dimension value. The edges in $E$ represent compatibility relationships between the vertices in $V$. Dimension values in the same dimension are considered incompatible $\forall k \in \{1, \dots, d\}, \forall v_i, v_j \in V_k, (v_i, v_j) \not\in E$
\item The number of testing configurations $n \in \mathbb{Z}^+$
\item Objective function $O(S)$, that when given a collection of testing configurations (``schedule") $S$, returns a non-negative number that quantifies how close the target distribution of dimension values is to the distribution of dimension values in $S$ (0 indicates a perfect match). 
\item Scope $C = (I = I_1 \cup \dots \cup I_d, X = X_1 \cup \dots \cup X_d)$, $\forall i, I_i \cap X_i = \emptyset$, which consists of collections of dimension values (for each dimension) that must be included in ($I$) and excluded from ($X$) the output.
\end{enumerate}

Output: 
\begin{itemize}
\item Schedule $S = [s_1, s_2, \dots, s_n]$
\end{itemize}

Constraints:
\begin{enumerate}
\item Schedule contains $n$ testing configurations, $|S| = n$
\item Each testing configuration has $d$ dimension values $\forall s \in S, |s| = d$, exactly one dimension value from each dimension $\forall s \in S, \forall i \in \{1, \dots, d\}, |s \cap V_i| = 1$
\item Dimension values in each testing configuration must all be compatible with each other $\forall s \in S, \forall s_i, s_j \in s,$ where $i \neq j$, $(s_i, s_j) \in E$
\item Schedule does not contain any dimension values in the exclude scope, $\forall s \in S, \forall i \in \{1, \dots, d\}, s \cap X_i = \emptyset$
\item Schedule covers every dimension value in the include scope, for dimensions where it is provided $\forall i \in \{1, \dots, d\}$ where $I_i \neq \emptyset, \forall v \in V_i \cap I_i, \exists s \in S, v \in s$
\item Schedule does not cover dimension values outside the include scope, for dimensions where it is provided, $\forall i \in \{1, \dots, d\}$ where $I_i \neq \emptyset, \forall v \in V_i - I_i, \nexists s \in S, v \in s$
\end{enumerate}

Optimize:
\begin{itemize}
\item Minimize the value of the objective function, $O(S)$
\item Optimal solution is $\text{min}_S \{O(S)\}$
\end{itemize}

There are 6 simulated annealing algorithms and 12 branch and bound algorithms developed with optimizn to solve the environment design problem, which (along with the environment design problem itself) are discussed in greater detail in other literature written by the same authors of this paper \cite{EnvDesign}.

\section{Experiments and Results}

Two experiments are run to evaluate the optimizn library. The first experiment tests out algorithms developed with optimizn on the traveling salesman problem. The second experiment tests out algorithms developed with optimizn on the environment design problem. 

\subsection{System Specifications}

Both experiments are run on their own Azure Databricks clusters, each with the following specifications.
\begin{itemize}
\itemsep 0em 
\item VMSKU: Standard\_DS3\_v2
\item Nodes: 1
\item Workers/Drivers: 1
\item Memory: 14 GB
\item Cores: 4
\item Databricks Runtime Version: 13.3 LTS
\end{itemize}

\subsection{Traveling Salesman Problem Experiment Design}
In the traveling salesman problem experiment, the following optimization algorithms will be tested.
\begin{itemize}
\itemsep 0em
\item optimizn, simulated annealing
\item python-tsp, simulated annealing \cite{PythonTSPSimAnneal} (an existing simulated annealing algorithm for the traveling salesman problem, from another library)
\item optimizn, traditional branch and bound (selection strategy: depth-first)
\item optimizn, look-ahead branch and bound (selection strategy: depth-first)
\item optimizn, traditional branch and bound (selection strategy: depth-first-best-first)
\item optimizn, look-ahead branch and bound (selection strategy: depth-first-best-first)
\item optimizn, traditional branch and bound (selection strategy: best-first-depth-first)
\item optimizn, look-ahead branch and bound (selection strategy: best-first-depth-first)
\end{itemize}

The optimizn algorithms are compared to the python-tsp simulated annealing algorithm to gauge how well the optimizn algorithms are performing with respect to an existing solver. python-tsp does have a branch and bound algorithm \cite{PythonTSPBnB}, but it is not included in the experiment since it cannot be run for a set period of time, cannot pick up from where previous runs left off, and does not accept an initial solution.

Between the optimizn and python-tsp simulated annealing algorithms, the method of generating a new solution from the current solution is the same, but the temperature values are computed differently. Additionally, the optimizn algorithm performs random restarts (with reset probability $\frac{1}{1500000}$) while the python-tsp algorithm does not.

The python-tsp algorithm does not keep track of the most optimal solution observed in each run. It only returns the current solution after a time limit is exceeded or when no improvement in solution optimality was observed after three iterations. For each of its runs, python-tsp algorithm is executed repeatedly (each execution starting from the solution of the previous one) until it has used up the compute time, and the most optimal solution returned across the executions is considered to be its most optimal solution for that run. Subsequent runs start from the most optimal solution observed in previous runs.

The experiment will be conducted on a graph of 200 cities. In this graph, each vertex represents a city, and each city's 2D-coordinates are drawn from a normal distribution with $\mu = 0$ and $\sigma = 5$. The weight of each edge is the Euclidean distance between its two cities. All algorithms start from the same initial solution, a path where the cities are in increasing, numerical order. Each algorithm will be run in a single stretch of time (three hours of compute time) and three times in succession (with one hour of compute time for each run), with the optimizn algorithms using continuous training and the python-tsp algorithm starting from the most optimal solution returned by its previous runs. The optimality of the solutions (path lengths) produced by each algorithm will be compared. 

\subsection{Traveling Salesman Problem Experiment Results} 

The costs of the most optimal solution (shortest path lengths) found by the algorithms are shown in Table \ref{TSPExpResultsScore}. For algorithms run in a single stretch, the results are shown only in the final column of the table.

``O-SA" and ``PT-SA" refer to the optimizn and python-tsp simulated annealing algorithms, respectively. ``TBnB" and ``LABnB" refer to the optimizn traditional and look-ahead branch and bound algorithms, respectively. The prefixes ``DF", ``DFBeF", and ``BeFDF" refer to branch and bound with the depth-first, depth-first-best-first, and best-first-depth-first selection strategies, respectively. The suffixes ``1" and ``2" represent the algorithm being run in a single stretch or in successive runs, respectively. 

\renewcommand{\arraystretch}{1.2}
\begin{table}[t]
\vskip 0in
\begin{center}
\begin{small}
\begin{sc}
\begin{tabular}{m{4em}|cccc}
\toprule
Opt. Alg. & Init. & Run 1 & Run 2 & Run 3 \\
\midrule
O-SA1 & 1765.716 & - & - & 369.981 \\ 
\hline
O-SA2 & 1765.716 & 373.806 & 353.297 & 353.297 \\
\hline
PT-SA1 & 1765.716 & - & - & 302.585 \\
\hline
PT-SA2 & 1765.716 & 305.806 & 299.140 & \textbf{291.383} \\
\hline
DF-LA-BnB1 & 1765.716 & - & - & 1642.857 \\
\hline
DF-LA-BnB2 & 1765.716 & 1630.178 & 1630.178 & 1630.178 \\
\hline
DFBeF-LA-BnB1 & 1765.716 & - & - & 311.041 \\
\hline
DFBeF-LA-BnB2 & 1765.716 & 311.041 & 311.041 & 311.041 \\
\hline
BeFDF-LA-BnB1 & 1765.716 & - & - & 1463.054 \\
\hline
BeFDF-LA-BnB2 & 1765.716 & 1550.719 & 1475.681 & 1475.681 \\
\hline
DF-T-BnB1 & 1765.716 & - & - & 1748.467 \\
\hline
DF-T-BnB2 & 1765.716 & 1748.467 & 1748.467 & 1748.467 \\
\hline
DFBeF-T-BnB1 & 1765.716 & - & - & 311.041 \\
\hline
DFBeF-T-BnB2 & 1765.716 & 311.041 & 311.041 & 311.041 \\
\hline
BeFDF-T-BnB1 & 1765.716 & - & - & 1765.716 \\
\hline
BeFDF-T-BnB2 & 1765.716 & 1765.716 & 1765.716 & 1765.716 \\
\bottomrule
\end{tabular}
\end{sc}
\end{small}
\end{center}
\caption{Lengths of shortest paths for the traveling salesman problem}
\vskip -0.1in
\label{TSPExpResultsScore}
\end{table}

\subsection{Environment Design Problem Experiment Design}

In the environment design problem experiment, simulated annealing and branch and bound algorithms developed using the optimizn library will be tested on the environment design problem for the Overlake scenario, a real testing scenario in Azure that pertains to new hardware models. This experiment and the tested optimization algorithms are discussed in greater detail in other literature written by the same authors of this paper \cite{EnvDesign}.

The algorithms will each be run three times in succession (with continuous training, one hour of compute time for each run) to improve the optimality of schedules (collections of testing configurations) for the Overlake scenario. The optimality of the schedules produced by each algorithm will be compared.

\subsection{Environment Design Problem Experiment Results}

The costs of the most optimal schedules produced by the algorithms are shown in Table \ref{EDExpResultsScore}.

The numbers from 1.1 to 1.6 correspond to the six simulated annealing algorithms, and the numbers from 2.1 to 2.6 and from 3.1 to 3.6 correspond to the twelve branch and bound algorithms.

\begin{table}[t]
\vskip 0in
\begin{center}
\begin{small}
\begin{sc}
\begin{tabular}{m{3em}|cccc}
\toprule
Opt. Alg. & Init. & Run 1 & Run 2 & Run 3 \\
\midrule
1.1 & 3.682e-2 & 2.162e-3 & 2.156e-3 & 2.156e-3 \\
\hline
1.2 & 3.682e-2 & \textbf{1.647e-3} & \textbf{1.647e-3} & \textbf{1.647e-3} \\
\hline
1.3 & 3.682e-2 & 1.347e-2 & 1.347e-2 & 1.347e-2 \\
\hline
1.4 & 3.682e-2 & 1.258e-2 & 1.258e-2 & 1.258e-2 \\
\hline
1.5 & 3.682e-2 & 2.239e-2 & 2.239e-2 & 2.239e-2 \\
\hline
1.6 & 3.682e-2 & 2.265e-2 & 2.265e-2 & 2.265e-2 \\
\hline
2.1 & 3.682e-2 & 3.682e-2 & 3.682e-2 & 3.682e-2 \\
\hline
2.2 & 3.682e-2 & 3.183e-2 & 3.183e-2 & 3.183e-2 \\
\hline
2.3 & 3.682e-2 & 3.604e-3 & 3.604e-3 & 3.604e-3 \\
\hline
2.4 & 3.682e-2 & 3.569e-3 & 3.568e-3 & 3.568e-3 \\
\hline
2.5 & 3.682e-2 & 3.682e-2 & 3.682e-2 & 3.682e-2 \\
\hline
2.6 & 3.682e-2 & 2.814e-2 & 2.767e-2 & 2.767e-2 \\
\hline
3.1 & 3.682e-2 & 3.505e-2 & 3.505e-2 & 3.505e-2 \\
\hline
3.2 & 3.682e-2 & 3.505e-2 & 3.505e-2 & 3.505e-2 \\
\hline
3.3 & 3.682e-2 & 3.694e-3 & 3.694e-3 & 3.694e-3 \\
\hline
3.4 & 3.682e-2 & 3.694e-3 & 3.694e-3 & 3.694e-3 \\
\hline
3.5 & 3.682e-2 & 3.682e-2 & 3.682e-2 & 3.682e-2 \\
\hline
3.6 & 3.682e-2 & 3.636e-2 & 3.601e-2 & 3.601e-2 \\
\bottomrule
\end{tabular}
\end{sc}
\end{small}
\end{center}
\caption{Objective function values of testing schedules for the environment design problem (Overlake scenario)}
\vskip -0.1in
\label{EDExpResultsScore}
\end{table}

\subsection{Analysis}

In the traveling salesman problem experiment, the best performing algorithm was the python-tsp simulated annealing algorithm, followed by the optimizn depth-first-best-first modified branch and bound algorithms (traditional and look-ahead). All algorithms saw significant improvement from the initial solution.

In the environment design problem experiment, the best performing algorithm was optimization algorithm 1.2, a simulated annealing algorithm. The \texttt{envdesign} model was able to produce schedules that were significantly more optimal than the initial expanded coverage schedule, indicating that the optimizn library is doing a good job helping the \texttt{envdesign} model optimize schedules.

The results of these experiments are a good indication that the optimizn library can be used to quickly and easily develop optimization algorithms to produce good solutions to a variety of optimization problems (well-known and niche). 

\section{Conclusion}
This paper presents optimizn, a Python library for developing customized optimization algorithms under general paradigms. optimizn's simulated annealing and branch and bound offerings make it quick, easy, and seamless to develop optimization algorithms. For both simulated annealing and branch and bound, optimizn supports continuous training, allowing the user to run their algorithms flexibly to obtain good solutions for complex and difficult optimization problems (well-known and niche).

The optimizn library has already been used to solve the environment design problem in Azure, and it can be used to solve other optimization problems and generate positive impact inside and outside of Azure. In the future, we intend to use the optimizn library to develop optimization algorithms to solve more niche, $NP$-hard optimization problems encountered in our work at Microsoft Azure, and improve optimizn based on the needs that arise in practice.

optimizn can be found on GitHub here: \url{https://github.com/microsoft/optimizn}, and on PyPI here: \url{https://pypi.org/project/optimizn/}. The code for the simulated annealing and branch and bound algorithms developed with optimizn for the traveling salesman problem can be found in optimizn's GitHub repository.

\section{Acknowledgements}
We would like to acknowledge Microsoft and Microsoft Azure for giving us the opportunity to build and showcase the optimizn library as well as apply it to our work at Microsoft Azure. We would also like to acknowledge the reviewers at Microsoft for their feedback and help during the paper's review/revision.

\newpage
\Urlmuskip=0mu plus 1mu\relax
\bibliographystyle{ieeetr}
\bibliography{Optimizn}

\begin{thebibliography}{10}

\bibitem{EnvDesign}
A.~Sathiya and R.~Pandey, ``The envdesign model: A method to solve the environment design problem,'' 2025.
\newblock \url{https://arxiv.org/abs/2412.18109}.

\bibitem{CLR4}
T.~H. Cormen, C.~E. Leiserson, R.~L. Rivest, and C.~Stein, {\em Introduction to Algorithms}.
\newblock Cambridge, Massachusetts: The MIT Press, 4~ed., 2022.
\newblock \url{https://a.co/d/eWV8aFt}.

\bibitem{SimAnnealOverview}
R.~A. Rutenbar, ``Simulated annealing algorithms: An overview,'' {\em IEEE Circuits and Devices Magazine}, vol.~5, pp.~19--26, January 1989.
\newblock \url{https://www.cs.amherst.edu/~ccmcgeoch/cs34/papers/rutenbar.pdf}. Online; accessed 8-January-2024.

\bibitem{BranchAndBound1}
J.~Clausen, ``Branch and bound algorithms - principles and examples..'' \url{https://imada.sdu.dk/u/jbj/heuristikker/TSPtext.pdf}, March 1999.
\newblock Online; accessed 16-December-2022.

\bibitem{BranchAndBound2}
A.~Bari, ``7.2 0/1 knapsack using branch and bound.'' \url{https://www.youtube.com/watch?v=yV1d-b_NeK8}, February 2018.
\newblock Online; accessed 16-December-2022.

\bibitem{OptimSAR}
T.~R. Foundation, ``Flexible optimization with simulated annealing.'' \url{https://search.r-project.org/CRAN/refmans/optimization/html/optim_sa.html}.
\newblock Online; accessed 3-January-2023.

\bibitem{SciPyAnneal}
SciPy, ``scipy.optimize.anneal.'' \url{https://docs.scipy.org/doc/scipy-0.14.0/reference/generated/scipy.optimize.anneal.html}, May 2014.
\newblock Online; accessed 3-January-2023.

\bibitem{SciPyDualAnnealing}
SciPy, ``scipy.optimize.dual\_annealing.'' \url{https://docs.scipy.org/doc/scipy/reference/generated/scipy.optimize.dual_annealing.html}, 2023.
\newblock Online; accessed 3-January-2023.

\bibitem{SciPyBasinHopping}
SciPy, ``scipy/scipy/optimize/\_basinhopping.py.'' \url{https://github.com/scipy/scipy/blob/890a1f2129313454e21280f5237ecb681f3127df/scipy/optimize/_basinhopping.py#L362}, December 2022.
\newblock Online; accessed 3-January-2023.

\bibitem{CPLEX}
IBM, ``Ibm® decision optimization cplex® modeling for python.'' \url{https://ibmdecisionoptimization.github.io/docplex-doc/}.
\newblock Online; accessed 24-February-2025.

\bibitem{PyBnB}
G.~A. Hackebeil, ``Welcome to pybnb - 0.6.2.'' \url{https://pybnb.readthedocs.io/en/stable/}.
\newblock Online; accessed 25-March-2023.

\bibitem{CPLEXBnB}
IBM, ``Tutorial: Beyond linear programming, (cplex part2).'' \url{https://ibmdecisionoptimization.github.io/tutorials/html/Beyond_Linear_Programming.html}.
\newblock Online; accessed 24-February-2025.

\bibitem{SimulatedAnnealing1}
T.~W. Schneider, ``The traveling salesman with simulated annealing, r, and shiny.'' \url{https://toddwschneider.com/posts/traveling-salesman-with-simulated-annealing-r-and-shiny/}, September 2014.
\newblock Online; accessed 8-January-2023.

\bibitem{SimulatedAnnealing2}
T.~W. Schneider, ``shiny-salesman/helpers.r.'' \url{https://github.com/toddwschneider/shiny-salesman/blob/master/helpers.R}, October 2014.
\newblock Online; accessed 8-January-2023.

\bibitem{PythonTSPSimAnneal}
F.~Goulart, T.~Frick, and Luan, ``fillipe-gsm/python-tsppython-tsp/python\_tsp/heuristics/simulated\_annealing.py.'' \url{https://github.com/fillipe-gsm/python-tsp/blob/master/python_tsp/heuristics/simulated_annealing.py}.
\newblock Online; accessed 27-March-20243.

\bibitem{PythonTSPBnB}
Luan, ``python-tsp/python\_tsp/exact/branch\_and\_bound/solver.py.'' \url{https://github.com/fillipe-gsm/python-tsp/blob/master/python_tsp/exact/branch_and_bound/solver.py}.
\newblock Online; accessed 7-March-2024.

\end{thebibliography}
\newpage

\end{document}